\documentclass[9pt,journal,cspaper,compsoc]{IEEEtran}

\usepackage{times}
\usepackage{epsfig}
\usepackage{graphicx}
\usepackage{amsmath}
\usepackage{amssymb}
\usepackage{color}
\usepackage{subfig}
\usepackage{tabularx}
\usepackage{multirow}
\usepackage[abs]{overpic}
\usepackage{algorithm}
\usepackage{algorithmic}
\usepackage{wrapfig}
\usepackage{ragged2e}
\usepackage[nocompress]{cite}
\usepackage[colorlinks,bookmarksnumbered,bookmarksopen,linkcolor=black,citecolor=black,urlcolor=black]{hyperref}

\usepackage{geometry}
\graphicspath{{./Imgs/}{./CVPR/Imgs/}}
\DeclareGraphicsExtensions{.pdf,.png,.jpg}
\newcolumntype{C}[1]{>{\centering\arraybackslash}p{#1}}
\newcommand{\tabincell}[2]{\begin{tabular}{@{}#1@{}}#2\end{tabular}}

\newcommand{\figref}[1]{Fig.~\ref{#1}}
\newcommand{\tabref}[1]{Fig.~\ref{#1}}
\newcommand{\secref}[1]{Sec.~\ref{#1}}

\def\ie{\emph{i.e.~}}
\def\eg{\emph{e.g.~}}
\def\etc{\emph{etc}}
\def\etal{{\em et al.}}
\def\conv{\emph{conv~}}
\def\sArt{{state-of-the-art~}}

\begin{document}
\title{Richer Convolutional Features for Edge Detection}

\author{Yun Liu, Ming-Ming Cheng, Xiaowei Hu, Jia-Wang Bian, Le Zhang,  
Xiang Bai, and Jinhui Tang \thanks{* M.M. Cheng is the corresponding author.
URL: http://mmcheng.net/rcfedge/}
\IEEEcompsocitemizethanks{
\IEEEcompsocthanksitem Y. Liu, M.M. Cheng, and J.W. Bian, are with
the College of Computer Science, Nankai University,
Tianjin 300350, China.
\IEEEcompsocthanksitem L. Zhang is with the Advanced Digital Sciences Center.
\IEEEcompsocthanksitem X. Bai is with Huazhong University of Science and Technology.
\IEEEcompsocthanksitem J. Tang is with School of Computer
   Science and Engineering, Nanjing University of Science and Technology, 
   Nanjing 210094, China.
\IEEEcompsocthanksitem A preliminary version of this work has been published in CVPR 2017 \cite{liu2017richer}.
}
}

\setcounter{page}{1939}
\markboth{IEEE TRANSACTIONS ON PATTERN ANALYSIS AND MACHINE INTELLIGENCE,~Vol.~41,No.~8, August 2019}%
{Liu \MakeLowercase{\textit{et al.}}: Richer Convolutional Features for Edge Detection}

\IEEEtitleabstractindextext{%
\begin{abstract}
\justifying
Edge detection is a fundamental problem in computer vision.
Recently, convolutional neural networks (CNNs) have pushed forward 
this field significantly.
Existing methods which adopt specific layers of deep CNNs may fail to capture 
complex data structures caused by variations of scales and aspect ratios.
In this paper, we propose an accurate edge detector using 
richer convolutional features (RCF). 
RCF encapsulates all convolutional features into more discriminative 
representation, which makes good usage of rich feature hierarchies, 
and is amenable to training via backpropagation.
RCF fully exploits multiscale and multilevel information of objects 
to perform the image-to-image prediction holistically.
Using VGG16 network, we achieve \sArt performance on several available datasets.
When evaluating on the well-known BSDS500 benchmark, we achieve ODS 
F-measure of \textbf{0.811} while retaining a fast speed (\textbf{8} FPS).
Besides, our fast version of RCF achieves ODS F-measure of \textbf{0.806} 
with \textbf{30} FPS.
We also demonstrate the versatility of the proposed method 
by applying RCF edges for classical image segmentation.
\end{abstract}

\begin{IEEEkeywords}
Edge detection, deep learning, richer convolutional features.
\end{IEEEkeywords}}

\maketitle

\IEEEdisplaynontitleabstractindextext
\IEEEpeerreviewmaketitle

\IEEEraisesectionheading{\section{Introduction}\label{sec:introduction}}
\IEEEPARstart{E}{dge} detection can be viewed as a method to extract 
visually salient edges and object boundaries from natural images. 
Due to its far-reaching applications in many high-level applications including 
object detection \cite{ullman1991recognition,ferrari2008groups},
object proposal generation \cite{zitnick2014edge,zhang2017sequential},
and image segmentation \cite{arbelaez2014multiscale,cheng2016hfs}, 
edge detection is a core low-level problem in computer vision.

The fundamental scientific question here is what is the appropriate 
representation which is rich enough for a predictor to distinguish 
edges/boundaries from the image data. 
To answer this, traditional methods first extract the local cues of brightness, 
color, gradient and texture, or other manually designed features like
Pb \cite{martin2004learning} and gPb \cite{arbelaez2011contour}, 
then sophisticated learning paradigms \cite{dollar2015fast}
are used to classify edge and non-edge pixels. 
Although low-level features based edge detectors are somehow promising,
their limitations are obvious as well.
For example, edges and boundaries are often defined to be semantically meaningful,
however, it is difficult to use low-level cues to represent high-level information.
Recently, \textit{convolutional neural networks} (CNNs) have become 
popular in computer vision \cite{simonyan2014very,long2015fully}.
Since CNNs have a strong capability to automatically learn the high-level 
representations for natural images, there is a recent trend of using CNNs 
to perform edge detection.
Some well-known CNN-based methods have pushed forward this field significantly,
such as DeepEdge \cite{bertasius2015deepedge}, N$^4$-Fields \cite{ganin2014n},
DeepContour \cite{shen2015deepcontour}, and HED \cite{xie2017holistically}.
Our algorithm falls into this category as well.

\newcommand{\AddImg}[2]{\subfloat[#1]{\includegraphics[width=.24\linewidth]{intro/289011#2}}}
\begin{figure}[!t]
  \centering
  \AddImg{original image}{} \hfill
  \AddImg{ground truth}{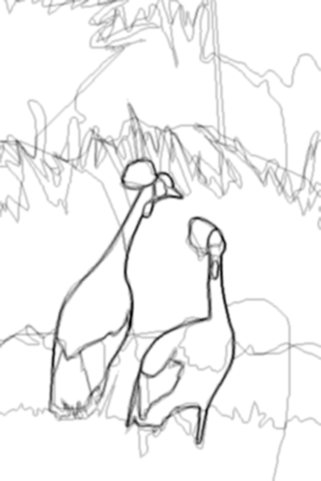} \hfill
  \AddImg{conv3\_1}{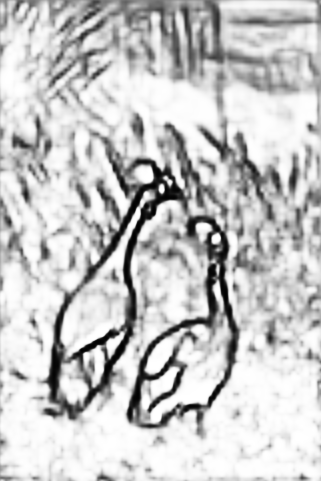} \hfill
  \AddImg{conv3\_2}{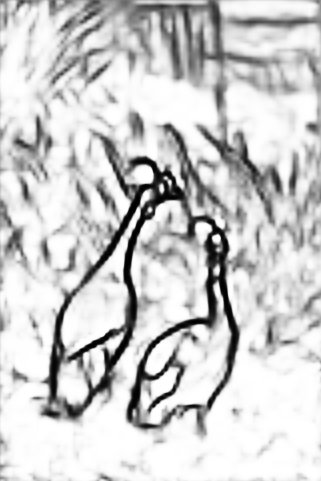} \\ \vspace{-0.15in}
  \AddImg{conv3\_3}{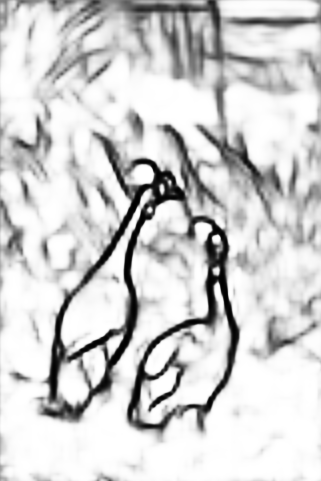} \hfill
  \AddImg{conv4\_1}{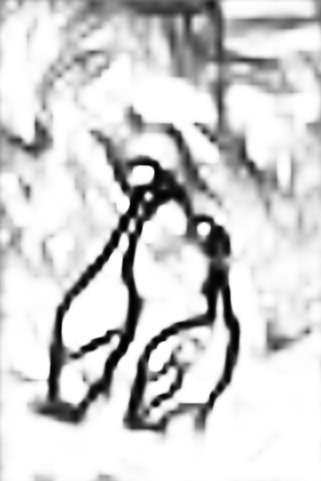} \hfill
  \AddImg{conv4\_2}{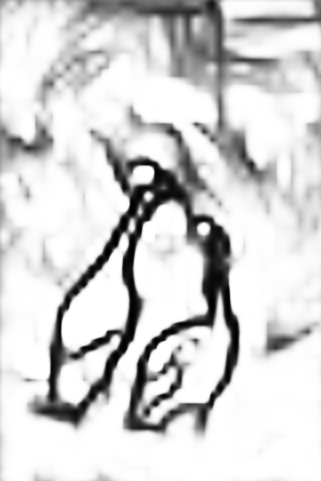} \hfill
  \AddImg{conv4\_3}{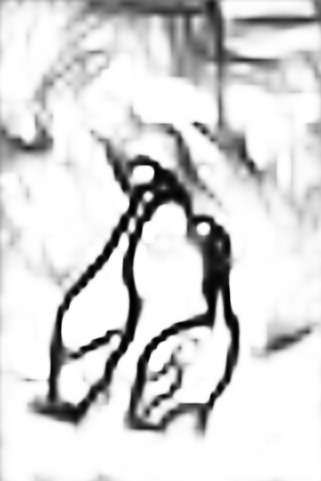} \\ \vspace{-0.1in}
  \caption{We build a simple network based on VGG16 \cite{simonyan2014very} 
 	to produce side outputs (c-h).
   	One can see that convolutional features become coarser gradually, 
    and the intermediate layers (c,d,f,g)  
    contain essential fine details that do not appear in other layers.
  }\label{fig:motivation}
  \vspace{-0.2in}
\end{figure}

As illustrated in \figref{fig:motivation}, we build a simple network to 
produce side outputs of intermediate layers using VGG16 \cite{simonyan2014very}
with HED architecture \cite{xie2017holistically}.
We can see that the information obtained by different \textit{convolution} 
(\ie \conv) layers gradually becomes coarser.
More importantly, intermediate \conv layers contain essential fine details.
However, previous CNN architectures only use the final \conv layer 
or the layers before the pooling layers of neural networks,
but ignore the intermediate layers.
On the other hand, since richer convolutional features are highly 
effective for many vision tasks, many researchers make efforts to 
develop deeper networks \cite{he2016deep}.
However, it is difficult to get the networks to converge when going 
deeper because of vanishing/exploding gradients and training data 
shortage (\eg for edge detection).
So why don't we make full use of the CNN features we have now?
Based on these observations, we propose richer convolutional 
features (RCF), a novel deep structure fully exploiting the CNN 
features from all the \conv layers, to perform the pixel-wise 
prediction for edge detection in an image-to-image fashion. 
RCF can automatically learn to combine complementary information 
from all layers of CNNs and thus can obtain accurate representations 
for objects or object parts in different scales.
The evaluation results demonstrate RCF performs very well on edge detection.

After the publication of the conference version \cite{liu2017richer}, 
our proposed RCF edges have been widely used in weakly supervised semantic
segmentation \cite{hou2018webseg}, style transfer \cite{liu2017depth}, 
and stereo matching \cite{song2018edgestereo}.
Besides, the idea of utilizing all the \textit{conv} layers in a unified 
framework can be potentially generalized to other vision tasks.
This has been demonstrated in skeleton detection \cite{zhao2018hi},
medial axis detection \cite{liu2017rsrn}, people detection \cite{zeng2017people},
and surface fatigue crack identification \cite{xu2018surface}.

When evaluating our method on BSDS500 dataset \cite{arbelaez2011contour} for 
edge detection, we achieve a good trade-off between effectiveness and efficiency
with the ODS F-measure of \textbf{0.811} and the speed of \textbf{8} FPS.
It even outperforms human perception (ODS F-measure 0.803).
In addition, a fast version of RCF is also presented,
which achieves ODS F-measure of \textbf{0.806} with \textbf{30} FPS.
When applying our RCF edges to classic image segmentation, we can obtain 
high-quality perceptual regions as well.

\section{Related Work}
As one of the most fundamental problem in computer vision, 
edge detection has been extensively studied for several decades.
\textit{Early pioneering methods} mainly focus on the utilization of intensity 
and color gradients, such as Canny \cite{canny1986computational}.
However, these early methods are usually not accurate enough 
for real-life applications.
To this end, \textit{feature learning based methods} have been proposed.
These methods, such as Pb \cite{martin2004learning}, 
gPb \cite{arbelaez2011contour}, and SE \cite{dollar2015fast},
usually employ sophisticated learning paradigms to predict edge strength 
with low-level features such as intensity, gradient, and texture. 
Although these methods are shown to be promising in some cases, 
these handcrafted features have limited ability to represent high-level 
information for semantically meaningful edge detection.

\textit{Deep learning based algorithms} have made vast inroads into 
many computer vision tasks. 
Under this umbrella, many deep edge detectors have been introduced recently.
Ganin \etal \cite{ganin2014n} proposed N$^4$-Fields that combines CNNs
with the nearest neighbor search.
Shen \etal \cite{shen2015deepcontour} partitioned contour data into subclasses
and fitted each subclass by learning the model parameters.
Recently, Xie \etal \cite{xie2017holistically} developed an efficient and 
accurate edge detector, HED, which performs image-to-image training and prediction.
This holistically-nested architecture connects their side output layers,
which is composed of one \textit{conv} layer with kernel size 1, 
one \textit{deconv} layer, and one softmax layer,
to the last \textit{conv} layer of each stage in VGG16 \cite{simonyan2014very}.
Moreover, Liu \etal \cite{liu2016learning} used relaxed labels generated by 
bottom-up edges to guide the training process of HED.
Wang \etal \cite{wang2018deep} leveraged a top-down backward refinement pathway
to effectively learn crisp boundaries.
Xu \etal \cite{xu2017learning} introduced a hierarchical deep model to 
robustly fuse the edge representations learned at different scales.
Yu \etal \cite{yu2017casenet} extended the success in edge detection
to semantic edge detection which simultaneously detected and recognized
the semantic categories of edge pixels.

Although these aforementioned CNN-based models have pushed the state of 
the arts to some extent, they all turn out to be lacking in our view 
because that they are not able to fully exploit the rich feature 
hierarchies from CNNs.
These methods usually adopt CNN features only from the last layer 
of each \textit{conv} stage.
To address this, we propose a fully convolutional network to combine 
features from all \conv layers efficiently.

\begin{figure}[tb]
    \centering
    \includegraphics[width=\linewidth]{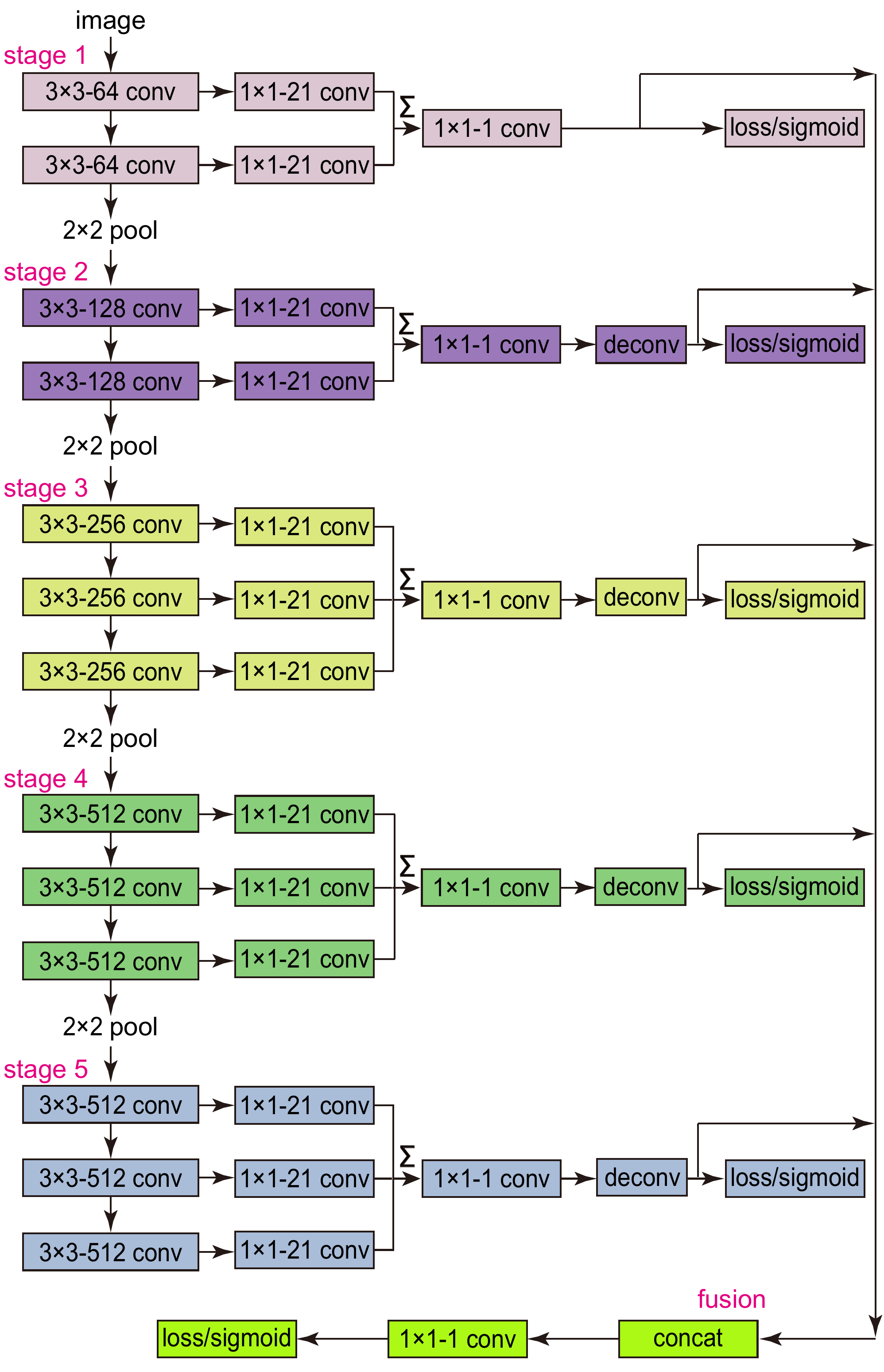} \\
    \caption{Our RCF network architecture. The input is an image with arbitrary sizes,
    	and our network outputs an edge possibility map in the same size.}
    \label{fig:netArchitecture}
\end{figure}

\section{Richer Convolutional Features (RCF)}
\subsection{Network Architecture}
We take inspirations from existing work \cite{long2015fully,xie2017holistically}
and embark on the VGG16 network \cite{simonyan2014very}.
VGG16 network composes of 13 \textit{conv} layers and 3 fully connected layers.
Its \textit{conv} layers are divided into five stages,
in which a pooling layer is connected after each stage.
The useful information captured by each \textit{conv} layer becomes coarser
with its receptive field size increasing.
Detailed receptive field sizes of different layers can be found 
in \cite{xie2017holistically}.
The use of this rich hierarchical information is hypothesized to help edge detection.
The starting point of our network design lies here.

The novel network introduced by us is shown in \figref{fig:netArchitecture}.
Compared with VGG16, our modifications can be summarized as following:
\begin{itemize}
\item We cut all the fully connected layers and the \textit{pool5} layer.
On the one side, we remove the fully connected layers to have 
a fully convolutional network for an image-to-image prediction.
On the other hand, adding \textit{pool5} layer will increase the stride by two times,
which usually leads to degeneration of edge localization.
\item Each \textit{conv} layer in VGG16 is connected to a \textit{conv} layer
with kernel size $1 \times 1$ and channel depth 21.
And the resulting feature maps in each stage are accumulated using 
an \textit{eltwise} layer to attain hybrid features.
\item An $1 \times 1-1$ \textit{conv} layer follows each \textit{eltwise} layer.
Then, a \textit{deconv} layer is used to up-sample this feature map.
\item A \textit{cross-entropy loss}/\textit{sigmoid} layer is connected to the up-sampling layer in each stage.
\item All the up-sampling layers are concatenated. 
Then an $1 \times 1$ \textit{conv} layer is used to fuse feature 
maps from each stage. 
At last, a \textit{cross-entropy loss}/\textit{sigmoid} layer is followed
to get the fusion loss/output.
\end{itemize}
In RCF, features from all \textit{conv} layers are well-encapsulated 
into a final representation in a holistic manner which is amenable to 
training by back-propagation. 
As receptive field sizes of \textit{conv} layers in VGG16 are different 
from each other, RCF endows a better mechanism than existing ones to 
learn multiscale information coming from all levels of convolutional 
features which we believe are all pertinent for edge detection.
In RCF, high-level features are coarser and can obtain strong response 
at the larger object or object part boundaries as illustrated 
in \figref{fig:motivation} while features from lower-part of CNNs 
are still beneficial in providing complementary fine details.


\subsection{Annotator-robust Loss Function}
Edge datasets in this community are usually labeled by several annotators
using their knowledge about the presence of objects or object parts.
Though humans vary in cognition, these human-labeled edges for 
the same image share high consistency \cite{martin2004learning}
For each image, we average all the ground truth to generate an edge 
probability map, which ranges from 0 to 1.
Here, 0 means no annotator labeled at this pixel, and 1 means all 
annotators have labeled at this pixel.
We consider the pixels with edge probabilities higher than $\eta$ as 
positive samples and the pixels with edge probabilities equal 
to 0 as negative samples.
Otherwise, if a pixel is marked by fewer than $\eta$ of the annotators,
this pixel may be semantically controversial to be an edge point.
Thus, regarding those pixels as either positive or negative samples 
may confuse the networks. 
Hence we ignore them, but HED tasks them as negative samples
and uses a fix $\eta$ of 0.5.

We compute the loss of each pixel with respect to its label as
\begin{equation} \label{eq:loss_pixel}
l(X_i; W) = \left\{
\begin{aligned}
\alpha \cdot log\ & (1-P(X_i; W)) \hspace{0.14in}  if\ y_i = 0 \\
0 & \hspace{1.07in} if\ 0 < y_i \leq \eta \\
\beta \cdot log\ & P(X_i; W) \hspace{0.48in}  otherwise, \\
\end{aligned}
\right.
\end{equation}
in which
\begin{equation} \label{eq:loss_coef}
\begin{aligned}
\alpha &= \lambda\cdot\frac{|Y^+|}{|Y^+|+|Y^-|} \\
\beta &= \frac{|Y^-|}{|Y^+|+|Y^-|}.
\end{aligned}
\end{equation}
$Y^+$ and $Y^-$ denote the positive sample set and the negative sample set,
respectively.
The hyper-parameter $\lambda$ is used to balance the number of 
positive and negative samples.
The activation value (CNN feature vector) and ground truth edge probability
at pixel $i$ are presented by $X_i$ and $y_i$, respectively.
$P(X)$ is the standard \textit{sigmoid} function,
and $W$ denotes all the parameters that will be learned in our architecture.
Therefore, our improved loss function can be formulated as
\begin{equation}
L(W) = \sum_{i=1}^{|I|} \Big(\sum_{k=1}^{K} l(X_i^{(k)}; W) + l(X_i^{fuse}; W)\Big),
\end{equation}
where $X_i^{(k)}$ is the activation value from stage $k$ while $X_i^{fuse}$ 
is from the fusion layer.
$|I|$ is the number of pixels in image $I$, and $K$ is the number of stages (equals to 5 here).

\subsection{Multiscale Hierarchical Edge Detection}
In single scale edge detection, we feed an original image into our 
fine-tuned RCF network, then, the output is an edge probability map.
To further improve the quality of edges,
we use image pyramids during the test phase.
Specifically, we resize an image to construct an image pyramid,
and each of these images is fed into our single-scale detector separately.
Then, all resulting edge probability maps are resized to 
the original image size using bilinear interpolation.
At last, these maps are fused to get the final prediction map.
We adopt simple average fusion in this study 
although other advanced strategies are also applicable.
In this way, our preliminary version \cite{liu2017richer} \textit{firstly} 
demonstrates multiscale testing is still beneficial for edge detection 
although RCF itself is able to encode multiscale information.
Considering the trade-off between accuracy and speed,
we use three scales 0.5, 1.0, and 1.5 in this paper.
When evaluating RCF on BSDS500 \cite{arbelaez2011contour} dataset,
this simple multiscale strategy improves the ODS F-measure 
from 0.806 to 0.811 with the speed of 8 FPS which we believe 
is good enough for real-life applications.
See \secref{sec:edge_eval_bsds} for details.

\subsection{Comparison With HED}
The most obvious differences between our RCF and HED \cite{xie2017holistically} 
lie in the three following aspects.

First, HED only considers the last \textit{conv} layer in each stage of VGG16,
in which lots of helpful information for edge detection is missed.
In contrast to it, RCF uses richer features from all the \textit{conv} layers,
making it more possible to capture more object or object-part boundaries 
across a larger range of scales.

Second, a novel loss function is proposed to treat training examples properly.
We consider the edge pixels that $\eta$ of the annotators labeled as positive samples
and the pixels that no annotator labeled as negative samples.
Besides, we ignore edge pixels that are marked by a few annotators 
because of their confusing attributes.
In contrast, HED view edge pixels that are marked by less than half of the
annotators as negative samples, which may confuse the network training
because these pixels are not true non-edge points.
Our new loss have been used in \cite{wang2018deep}.

Thirdly, our preliminary version \cite{liu2017richer} first proposes 
the multiscale test for edge detection.
Recent edge detectors such as HED usually use multiscale network features, 
but we demonstrate the simple multiscale test is still helpful
to edge detection.
This idea is also accepted by recent work \cite{wang2018deep}.


\begin{figure*}[!t]
    \centering
    \subfloat[]{
    	\includegraphics[height=.4\linewidth,width=.45\linewidth]{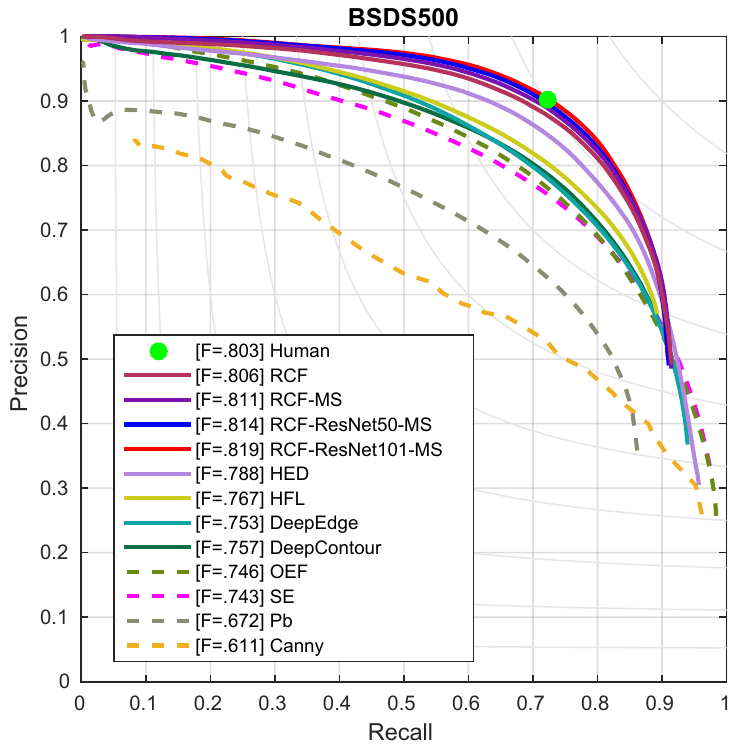}
        \label{fig:bsds_eval}}
    \hspace{0.16in}
    \subfloat[]{
    	\includegraphics[height=.4\linewidth,width=.45\linewidth]{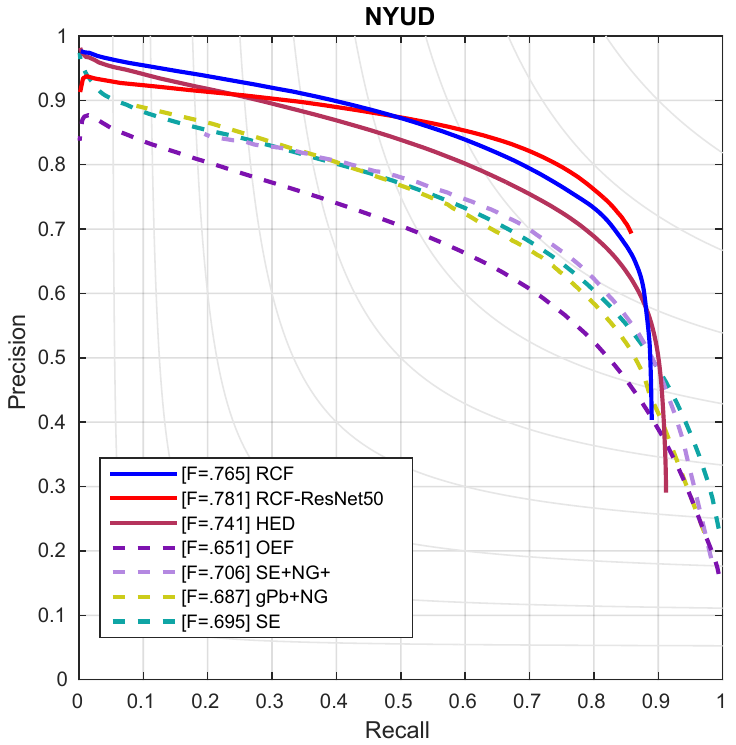}
        \label{fig:nyud_eval}}
    \vspace{-.1in}
    \caption{The evaluation results on the standard BSDS500 \cite{arbelaez2011contour} 
    and NYUD \cite{silberman2012indoor} datasets. The multiscale version of RCF is only
    evaluated on the BSDS500 dataset. Here, the solid lines represent CNN based
    methods, while the dotted lines represent non-deep algorithms.}
    \vspace{-0.1in}
\end{figure*}

\section{Experiments on Edge Detection} \label{sec:edge_experiments}
%
We implement our network using the Caffe framework \cite{jia2014caffe}.
The default setting using VGG16 \cite{simonyan2014very} backbone net, 
and we also test ResNet \cite{he2016deep} backbone net.
In RCF training, the weights of $1 \times 1$ \textit{conv} layers in stage 1-5 
are initialized from zero-mean Gaussian distributions with standard deviation 
0.01 and the biases are initialized to 0.
The weights of the $1 \times 1$ \textit{conv} layer in the fusion stage are 
initialized to 0.2 and the biases are initialized to 0.
The weights of other layers are initialized using pre-trained ImageNet models.
Stochastic gradient descent (SGD) minibatch samples 10 images randomly 
in each iteration.
For other SGD hyper-parameters, the global learning rate is set to 1e-6
and will be divided by 10 after every 10k iterations.
The momentum and weight decay are set to 0.9 and 0.0002, respectively.
We run SGD for 40k iterations totally.
The parameters $\eta$ and $\lambda$ in loss function are set 
depending on the training data.
All experiments in this paper are finished using a NVIDIA TITAN X GPU.

Given an edge probability map, a threshold is needed to produce the binary edge map.
There are two choices to set this threshold.
The first one is referred as \textit{optimal dataset scale} (ODS) 
which employs a fixed threshold for all images in a dataset.
The second is called \textit{optimal image scale} (OIS) 
which selects an optimal threshold for each image.
We report the F-measure ($\frac{2 \cdot Precision \cdot Recall}{Precision+Recall}$) 
of both ODS and OIS in our experiments.

\subsection{BSDS500 Dataset} \label{sec:edge_eval_bsds}
BSDS500 \cite{arbelaez2011contour} is a widely used dataset in edge detection.
It is composed of 200 training, 100 validation and 200 test images,
each of which is labeled by 4 to 9 annotators.
We utilize the training and validation sets for fine-tuning, and test set for evaluation.
Data augmentation is the same as \cite{xie2017holistically}.
Inspired by the previous work \cite{liu2016learning,yang2016object,kokkinos2015pushing}, we mix the augmented data
of BSDS500 with flipped VOC Context dataset \cite{mottaghi2014role} as training data.
When training, we set loss parameters $\eta$ and $\lambda$ to 0.5 and 1.1, respectively.
When evaluating, standard non-maximum suppression (NMS) \cite{dollar2015fast} is applied to thin detected edges.
We compare our method with some non-deep-learning algorithms, including
Canny \cite{canny1986computational}, Pb \cite{martin2004learning}, 
SE \cite{dollar2015fast}, and OEF \cite{hallman2015oriented},
and some recent deep learning based approaches, including
DeepContour \cite{shen2015deepcontour}, DeepEdge \cite{bertasius2015deepedge},
HED \cite{xie2017holistically}, HFL \cite{bertasius2015high},
MIL+G-DSN+MS+NCuts \cite{kokkinos2015pushing},
CASENet \cite{yu2017casenet}, AMH \cite{xu2017learning},
CED \cite{wang2018deep} and \etc.

\figref{fig:bsds_eval} shows the evaluation results.
The performance of human eye in edge detection is known as 0.803 ODS F-measure.
Both single-scale and multiscale (MS) versions of RCF get better results 
than average human performance.
When comparing with HED \cite{xie2017holistically}, ODS F-measures of our 
RCF-MS and RCF are 2.3\% and 1.8\% higher than it, respectively.
Moreover, ResNet50 and ResNet101 can further improve the performance
with more \textit{conv} layers.
These results demonstrate the effectiveness of the richer convolutional features.

\begin{figure}[!ht]
    \centering
    \begin{tabular}{c|c|c|c} \hline
    	Method & ODS & OIS & FPS \\ \hline
    	Canny \cite{canny1986computational} & 0.611 & 0.676 & 28 \\
    	Pb \cite{martin2004learning} & 0.672 & 0.695 & - \\
    	SE \cite{dollar2015fast} & 0.743 & 0.763 & 2.5 \\
    	OEF \cite{hallman2015oriented} & 0.746 & 0.770 & 2/3 \\ \hline
    	DeepContour \cite{shen2015deepcontour} & 0.757 & 0.776 & 1/30$^\dag$ \\
    	DeepEdge \cite{bertasius2015deepedge} & 0.753 & 0.772 & 1/1000$^\dag$ \\
    	HFL \cite{bertasius2015high} & 0.767 & 0.788 & 5/6$^\dag$ \\
    	N$^4$-Fields \cite{ganin2014n} & 0.753 & 0.769 & 1/6$^\dag$ \\
    	HED \cite{xie2017holistically} & 0.788 & 0.808 & \bf{30}$^\dag$ \\
    	RDS \cite{liu2016learning} & 0.792 & 0.810 & \bf{30}$^\dag$ \\
    	CEDN \cite{yang2016object} & 0.788 & 0.804 & 10$^\dag$ \\
    	\tabincell{c}{MIL+G-DSN+VOC+MS\\+NCuts \cite{kokkinos2015pushing}} 
        	& 0.813 & 0.831 & 1$^\dag$ \\ 
        CASENet \cite{yu2017casenet} & 0.767 & 0.784 & 18$^\dag$ \\
        AMH-ResNet50 \cite{xu2017learning} & 0.798 & 0.829 & - \\ 
        CED-VGG16 \cite{wang2018deep} & 0.794 & 0.811 & - \\ 
        CED-ResNet50+VOC+MS \cite{wang2018deep} & 0.817 & 0.834 & - \\ \hline
    	RCF & 0.806 & 0.823 & \bf{30}$^\dag$ \\
    	RCF-MS & 0.811 & 0.830 & 8$^\dag$ \\ 
        RCF-ResNet50 & 0.808 & 0.825 & 20$^\dag$ \\
        RCF-ResNet50-MS & 0.814 & 0.833 & 5.4$^\dag$ \\
        RCF-ResNet101 & 0.812 & 0.829 & 12.2$^\dag$ \\
        RCF-ResNet101-MS & \bf{0.819} & \bf{0.836} & 3.6$^\dag$ \\ \hline
    \end{tabular}
    \caption{The comparison with some competitors on the 
    BSDS500 \cite{arbelaez2011contour} dataset. \dag~means GPU time.}
    \label{tab:bsds_eval}
    \vspace{-0.1in}
\end{figure}

We show statistic comparison in \tabref{tab:bsds_eval}.
From RCF to RCF-MS, the ODS F-measure increases from 0.806 to 0.811,
though the speed drops from 30 FPS to 8 FPS.
It proves the validity of our multiscale strategy.
RCF with ResNet101 \cite{he2016deep} achieves a \sArt 0.819 ODS F-measure.
We also observe an interesting phenomenon in which the RCF curves 
are not as long as other methods when evaluated using the default 
parameters in BSDS500 benchmark.
It may suggest that RCF tends to only remain very confident edges.
%
%
Our methods also achieve better results than recent edge detectors,
such as AMH \cite{xu2017learning} and CED \cite{wang2018deep}.
Note that AMH and CED use complex networks with more 
weights than our simple RCF.
Our RCF network only adds some $1 \times 1$ \textit{conv} layers to HED,
so the time consumption is on par with HED.
We can see that RCF achieves a good trade-off between effectiveness and efficiency.


\subsection{NYUD Dataset}
NYUD \cite{silberman2012indoor} dataset is composed of 1449 densely labeled pairs
of aligned RGB and depth images captured from indoor scenes.
Recently, many works have conducted edge evaluation on it, such as \cite{dollar2015fast}.
Gupta \etal \cite{gupta2013perceptual} split NYUD dataset into 381 training, 414 validation, and 654 test images.
We follow their settings and train RCF using the training
and validation sets as in HED \cite{xie2017holistically}.

We utilize depth information by using HHA \cite{gupta2014learning},
in which depth information is encoded into three channels:
horizontal disparity, height above ground, and angle with gravity.
Thus HHA features can be represented as a color image by normalization.
Then, two models for RGB images and HHA feature images are trained separately.
In the training process, $\lambda$ is set to 1.2 for both RGB and HHA.
Since NYUD only has one ground truth for each image, $\eta$ is useless here.
Other network settings are the same as used for BSDS500.
At the test phase, the final edge predictions are defined 
by averaging the outputs of RGB model and HHA model.
Since there is already an average operation, 
the multiscale test is not evaluated here. 
When evaluating, we increase localization tolerance, which controls the 
maximum allowed distance in matches between predicted edges and ground truth, 
from 0.0075 to 0.011, because images in NYUD dataset are larger than images 
in BSDS500 dataset.

\begin{figure}[t]
    \centering
    \begin{tabular}{c||c|c|c} \hline
    	Method & ODS & OIS & FPS \\ \hline \hline
    	OEF \cite{hallman2015oriented} & 0.651 & 0.667 & 1/2 \\
    	gPb+NG \cite{gupta2013perceptual}  & 0.687 & 0.716 & 1/375 \\
    	SE \cite{dollar2015fast} & 0.695 & 0.708 & 5 \\
    	SE+NG+ \cite{gupta2014learning} & 0.706 & 0.734 & 1/15 \\ \hline
    	HED-HHA \cite{xie2017holistically} & 0.681 & 0.695 
        	& \textbf{20}$^\dag$ \\
    	HED-RGB \cite{xie2017holistically} & 0.717 & 0.732 
        	& \textbf{20}$^\dag$ \\
    	HED-RGB-HHA \cite{xie2017holistically} & 0.741 
        	& 0.757 & 10$^\dag$ \\ \hline
    	RCF-HHA & 0.703 & 0.717 & \textbf{20}$^\dag$ \\
    	RCF-RGB & 0.743 & 0.757 & \textbf{20}$^\dag$ \\
    	RCF-RGB-HHA & 0.765 & 0.780 & 10$^\dag$ \\ 
        RCF-ResNet50-RGB-HHA & \textbf{0.781} & \textbf{0.793} & 7$^\dag$ \\
        \hline
    \end{tabular}
    \caption{The comparison with some competitors on the 
    	NYUD dataset \cite{silberman2012indoor}. \dag means GPU time.}
    \label{tab:nyud_eval}
    \vspace{-0.1in}
\end{figure}

\renewcommand{\AddImg}[1]{\includegraphics[width=.245\linewidth]{samples/#1}}
\newcommand{\AddImgs}[1]{\AddImg{#1-img} \AddImg{#1-gt} \AddImg{#1-rcf} \AddImg{#1-ucm} \\ \vspace{.02in} }
\begin{figure*}[!t]
    \centering
    \AddImgs{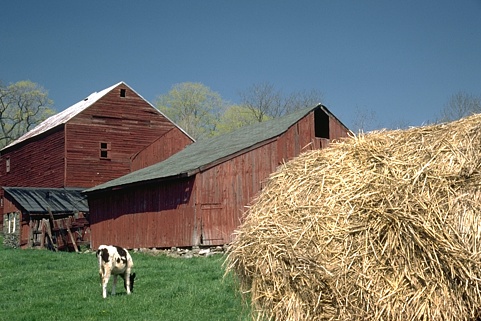}
    \AddImgs{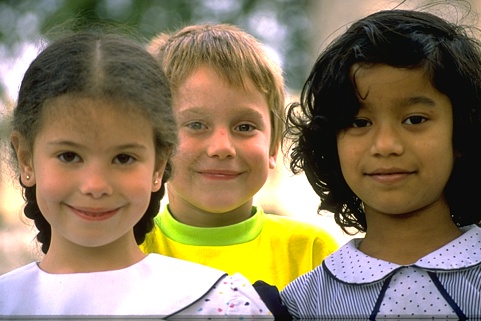}
    \AddImgs{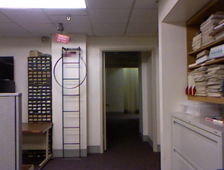}
    \AddImgs{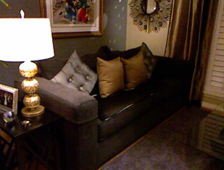}
    \caption{Some examples of RCF. 
        \textbf{Top two rows}: BSDS500 \cite{arbelaez2011contour}.
    	\textbf{Bottom two rows}: NYUD \cite{silberman2012indoor}.
        From \textbf{Left} to \textbf{Right}: origin image, ground truth, 
        RCF edge map, RCF UCM map.}
    \label{fig:samples}
\end{figure*}

We compare our single-scale version of RCF with some well-established competitors.
OEF \cite{hallman2015oriented} only uses RGB images,
while other methods employ both depth and RGB information.
The precision-recall curves are shown in \figref{fig:nyud_eval}.
RCF achieves competitive performance on NYUD dataset,
and it is significantly better than HED.
\tabref{tab:nyud_eval} shows the statistical comparison.
We can see that RCF outperforms HED not only 
on separate HHA or RGB data, but also on the merged RGB-HHA data.
For HHA and RGB data, ODS F-measure of RCF is 2.2\% and 2.6\% higher 
than HED, respectively.
For merging RGB-HHA data, RCF is 2.4\% higher than HED.
Furthermore, HHA edges perform worse than RGB,
but averaging HHA and RGB edges achieves much higher results.
It suggests that combining different types of information is very useful 
for edge detection, and this may explain why OEF performs much 
worse than other methods.
RCF with ResNet50 \cite{he2016deep} improves a 1.6\% ODS F-measure 
when compared with RCF with VGG16 \cite{simonyan2014very}.

\begin{figure*}[t]
	\centering
    \includegraphics[height=.4\linewidth,width=.45\linewidth]{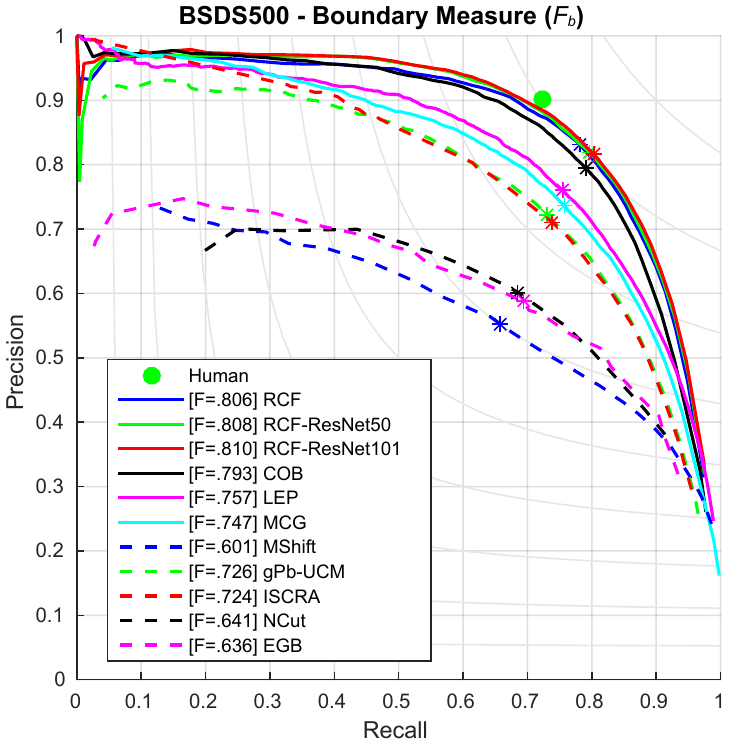}
    \hspace{0.2in}
    \includegraphics[height=.4\linewidth,width=.45\linewidth]{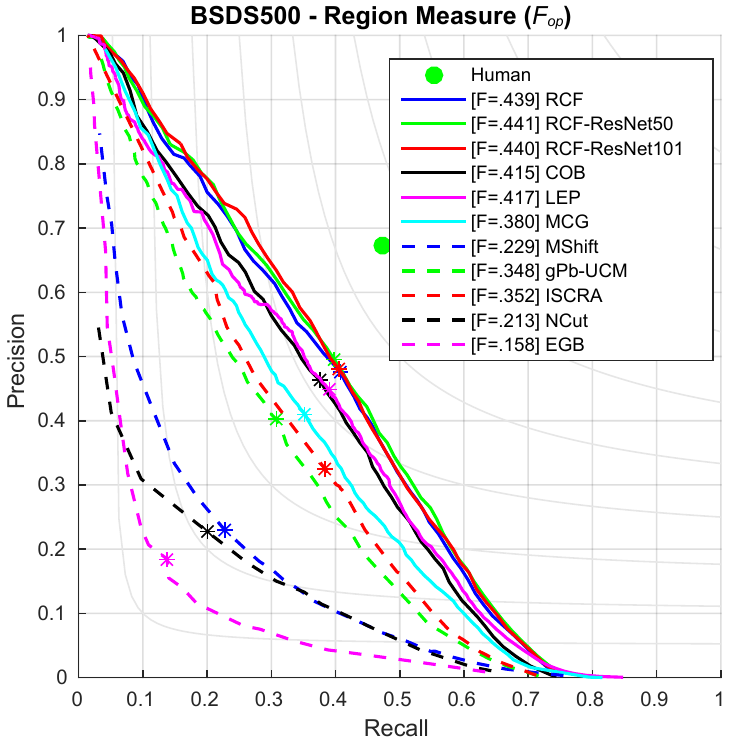}
    \caption{The precision-recall curves for the evaluation of boundary measure
    	($F_b$ \cite{martin2004learning}) and region measure 
        ($F_{op}$ \cite{pont2016supervised}) of classical image 
        segmentation on the BSDS500 test set \cite{arbelaez2011contour}.}
    \label{fig:img_seg_bsds500}
\end{figure*}

\begin{figure*}[t]
	\centering
    \includegraphics[height=.4\linewidth,width=.45\linewidth]{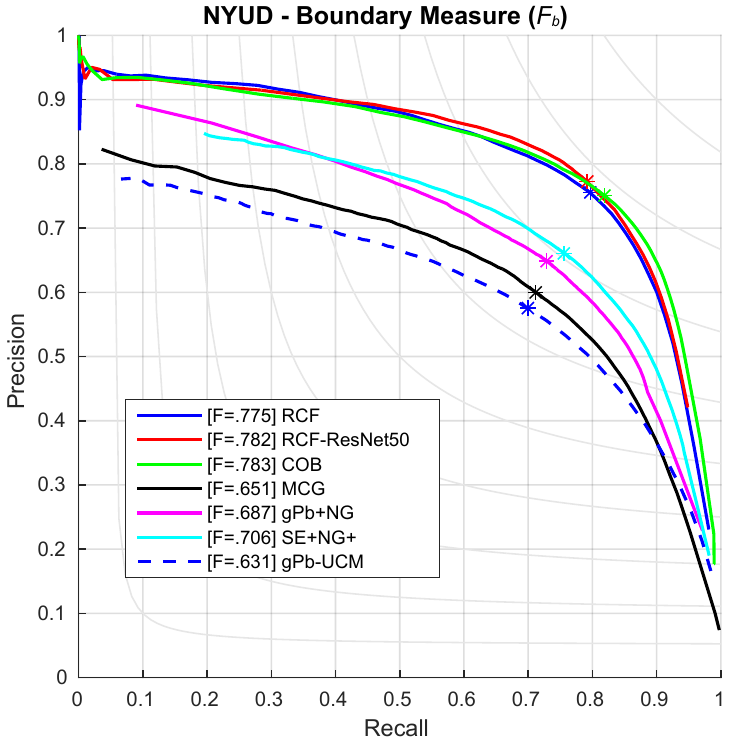}
    \hspace{0.2in}
    \includegraphics[height=.4\linewidth,width=.45\linewidth]{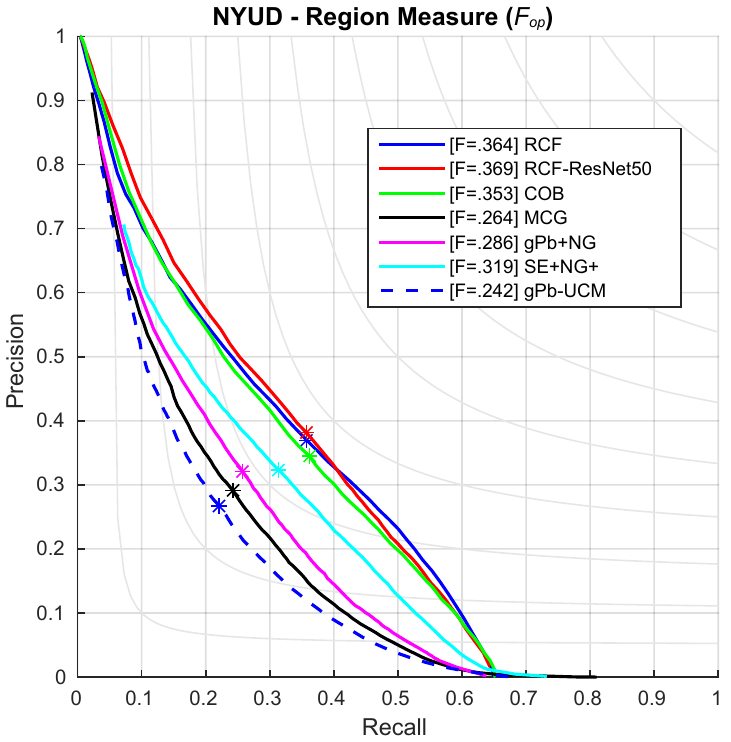}
    \caption{The precision-recall curves for the evaluation of boundary measure
    	($F_b$ \cite{martin2004learning}) and region measure 
        ($F_{op}$ \cite{pont2016supervised}) of classical image 
        segmentation on the NYUD test set \cite{silberman2012indoor}.}
    \label{fig:img_seg_nyud}
\end{figure*}

\subsection{Multicue Dataset}
Multicue dataset is proposed by M{\'e}ly \etal \cite{mely2016systematic}
to study psychophysics theory for boundary detection.
It is composed of short binocular video sequences of 100 challenging natural scenes
captured by a stereo camera.
Each scene contains a left and a right view short (10-frame) color sequences.
The last frame of the left images for each scene is labeled for two annotations:
object boundaries and low-level edges.
Unlike people who usually use boundary and edge interchangeably,
they strictly defined boundary and edge according to visual perception 
at different stages.
Thus, boundaries are referred to the boundary pixels of meaningful objects,
and edges are abrupt pixels at which the luminance, color, or stereo changes sharply.
In this subsection, we use boundary and edge as defined by 
M{\'e}ly \etal \cite{mely2016systematic} while considering boundary 
and edge having the same meaning in previous sections.

As done in M{\'e}ly \etal \cite{mely2016systematic} and HED \cite{xie2017holistically},
we randomly split these human-labeled images into 80 training and 20 test samples,
and average the scores of three independent trials as final results.
When training on Multicue, $\lambda$ is set to 1.1,
and $\eta$ is set to 0.4 for boundary task and 0.3 for edge task.
For boundary detection task, we use learning rate 1e-6 and run SGD for 2k iterations.
For edge detection task, we use learning rate 1e-7 and run SGD for 4k iterations.
Since the image resolution of Multicue is very high, 
we randomly crop $500 \times 500$ patches from original images at each iteration.

\begin{figure}[t]
    \centering
    \begin{tabular}{c||c|c} \hline
    	Method & ODS & OIS \\ \hline \hline
    	Human-Boundary \cite{mely2016systematic} & 0.760 (0.017) & -- \\
        Multicue-Boundary \cite{mely2016systematic} & 0.720 (0.014) & -- \\
        HED-Boundary \cite{xie2017holistically} 
        	& 0.814 (0.011) & 0.822 (0.008) \\
        RCF-Boundary & 0.817 (\textbf{0.004}) & 0.825 (\textbf{0.005}) \\
        RCF-MS-Boundary & \textbf{0.825} (0.008) & \textbf{0.836} (0.007) \\ \hline
        Human-Edge \cite{mely2016systematic} & 0.750 (0.024) & -- \\
        Multicue-Edge \cite{mely2016systematic} & 0.830 (\textbf{0.002}) & -- \\
        HED-Edge \cite{xie2017holistically} & 0.851 (0.014) 
        	& \textbf{0.864} (0.011) \\
        RCF-Edge & 0.857 (0.004) & 0.862 (\textbf{0.004}) \\
        RCF-MS-Edge & \textbf{0.860} (0.005) & \textbf{0.864} (\textbf{0.004}) \\
        \hline
    \end{tabular}
    \caption{The comparisons on the Multicue dataset \cite{mely2016systematic}. 
    The numbers in the parentheses mean standard deviations.}
    \label{tab:multicue_eval}
    \vspace{-0.1in}
\end{figure}

We use VGG16 \cite{simonyan2014very} as the backbone net.
The evaluation results are summarized in \tabref{tab:multicue_eval}.
Our proposed RCF achieves substantially higher results than HED.
For boundary task, RCF-MS is 1.1\% ODS F-measure higher and 1.4\% OIS F-measure higher than HED.
For edge task, RCF-MS is 0.9\% ODS F-measure higher than HED.
Note that the fluctuation of RCF is also smaller than HED,
which suggests RCF is more robust over different kinds of images.
Some qualitative results are shown in \figref{fig:samples}.

\subsection{Network Discussion}
To further explore the effectiveness of our network architecture,
we implement some mixed networks using VGG16 \cite{simonyan2014very}
by connecting our richer feature side outputs to some convolution stages
while connecting side outputs of HED to the other stages.
With training only on BSDS500 \cite{arbelaez2011contour} dataset and testing on the single scale,
evaluation results of these mixed networks are shown in \tabref{tab:discuss}.
The last two lines of this table correspond to HED and RCF, respectively.
We can observe that all of these mixed networks perform better than HED and worse than RCF
that is fully connected to RCF side outputs.
It clearly demonstrates the importance of our strategy of richer convolutional features.

\begin{figure}[!ht]
    \centering
    \begin{tabular}{c|c||c|c} \hline
    	RCF Stage & HED Stage & ODS & OIS \\ \hline
    	1, 2 & 3, 4, 5 & 0.792 & 0.810 \\
        2, 4 & 1, 3, 5 & 0.795 & 0.812 \\
        4, 5 & 1, 2, 3 & 0.790 & 0.810 \\
        1, 3, 5 & 2, 4 & 0.794 & 0.810 \\
        3, 4, 5 & 1, 2 & 0.796 & 0.812 \\ \hline
        -- & 1, 2, 3, 4, 5 & 0.788 & 0.808 \\
        1, 2, 3, 4, 5 & -- & \textbf{0.798} & \textbf{0.815} \\ \hline
    \end{tabular}
    \caption{Results of some thought networks.}
    \label{tab:discuss}
    \vspace{-0.15in}
\end{figure}

In order to investigate whether including additional nonlinearity helps,
we connecting ReLU layer after $1 \times 1 - 21$ or  $1 \times 1 - 1$ \textit{conv} layers in each stage.
However, the network performs worse.
Especially, when we attempt to add nonlinear layers to  $1 \times 1 - 1$ \textit{conv} layers,
the network can not converge properly.

\begin{figure}[t]
    \centering
    \begin{tabular}{c||p{0.3in}<{\centering}|p{0.3in}<{\centering}|%
    	p{0.3in}<{\centering}|p{0.3in}<{\centering}} \hline
    	\multirow{2}*{Methods} & \multicolumn{2}{c|}{Boundaries ($F_b$)} 
        	& \multicolumn{2}{c}{Regions ($F_{op}$)} \\ \cline{2-5}
        	& ODS & OIS & ODS & OIS \\ \hline
        NCut \cite{shi2000normalized} & 0.641 & 0.674 & 0.213 & 0.270 \\
        MShift \cite{comaniciu2002mean} & 0.601 & 0.644 & 0.229 & 0.292 \\
        EGB \cite{felzenszwalb2004efficient} & 0.636 & 0.674 & 0.158 & 0.240 \\
        gPb-UCM \cite{arbelaez2011contour} & 0.726 & 0.760 & 0.348 & 0.385 \\
        ISCRA \cite{ren2013image} & 0.724 & 0.752 & 0.352 & 0.418 \\
        MCG \cite{arbelaez2014multiscale} & 0.747 & 0.779 & 0.380 & 0.433 \\
        LEP \cite{zhao2015segmenting} & 0.757 & 0.793 & 0.417 & 0.468 \\
        COB \cite{maninis2016convolutional} & 0.793 & 0.820 & 0.415 & 0.466 \\ \hline
        RCF & 0.806 & 0.833 & 0.439 & 0.496 \\
        RCF-ResNet50 & 0.808 & 0.833 & \textbf{0.441} & 0.500 \\
        RCF-ResNet101 & \textbf{0.810} & \textbf{0.836} & 0.440 & \textbf{0.501} \\
        \hline
    \end{tabular}
    \caption{Evaluation results of boundaries ($F_b$ \cite{martin2004learning})
    and regions ($F_{op}$ \cite{pont2016supervised}) on the 
    BSDS500 test set \cite{arbelaez2011contour}.}
    \label{tab:img_seg_bsds500}
    \vspace{-0.15in}
\end{figure}

\section{Experiments on Image Segmentation}
The predicted edges of natural images are often used in another low-level vision 
technique, image segmentation, which aims to cluster similar pixels to form 
perceptual regions.
To transform a predicted edge map into a segmentation, Arbel{\'a}ez
\cite{arbelaez2011contour} introduced the Ultrametric Contour Map (UCM) that
can generate different image partitions when thresholding this hierarchical 
contour map at various contour probability values.
MCG \cite{arbelaez2014multiscale} develops a fast normalized cuts algorithm 
to accelerate \cite{arbelaez2011contour} and makes effective use of
multiscale information to generate an accurate hierarchical segmentation tree.
Note that MCG needs edge orientations as input.
These orientations are usually computed using simple morphological operations.
COB \cite{maninis2016convolutional} simultaneously predicts the magnitudes and
orientations of edges using HED-based CNNs, and then applies MCG framework to convert
these predictions to UCM.
Since much more accurate edge orientations are used, COB achieves the \sArt segmentation
results.

In order to demonstrate the versatility of the proposed method, 
here we evaluate the edges of RCF in the context of image segmentation.
Specifically, we apply the COB framework but replacing the 
HED edges with our RCF edges to perform image segmentation.
We evaluate the resulting segmenter on the
BSDS500 \cite{arbelaez2011contour} and NYUD \cite{silberman2012indoor} datasets.
Note that COB uses ResNet50 as its backbone net, so we also test
RCF with ResNet for fair comparison.
Besides the boundary measure ($F_b$) \cite{martin2004learning} used in 
\secref{sec:edge_experiments}, we also use the evaluation metric of precision-recall 
for objects and parts ($F_{op}$) \cite{pont2016supervised} to 
evaluate the region similarity between the segmentation and the corresponding ground truth.

\subsection*{BSDS500 Dataset}
On the challenging BSDS500 dataset \cite{arbelaez2011contour}, we compare RCF with 
some well-known generic image segmenters, including NCut \cite{shi2000normalized},
MShift \cite{comaniciu2002mean}, EGB \cite{felzenszwalb2004efficient},
gPb-UCM \cite{arbelaez2011contour}, ISCRA \cite{ren2013image},
MCG \cite{arbelaez2014multiscale}, LEP \cite{zhao2015segmenting},
and COB \cite{maninis2016convolutional}.
The evaluation results are shown in \figref{fig:img_seg_bsds500}.
RCF achieves the new state of the art on this dataset, 
both in terms of boundary and region quality.
COB \cite{maninis2016convolutional} gets the second place.
We show the numeric comparison in \tabref{tab:img_seg_bsds500}.
For the boundary measure, both the ODS and OIS F-measure of RCF are 1.3\% 
higher than COB.
For the region measure, the ODS and OIS F-measure of RCF are 2.4\%
and 3.0\% higher than COB, respectively.
Using the ResNet as the backbone net, RCF can further improve performance.
Since COB uses the edges produced by HED \cite{xie2017holistically}, 
these results demonstrate the effectiveness of RCF architecture.
From \figref{fig:img_seg_bsds500}, we can also see that although the 
boundary measure of RCF segmentation have reached human performance, 
the region measure is still far away from the human performance.
It indicates that better perception regions should be the main pursuit 
of classical image segmentation in the future.

\subsection*{NYUD Dataset}
On the RGB-D dataset of NYUD \cite{silberman2012indoor}, we compare not only with
some RGB based methods, \eg gPb-UCM \cite{arbelaez2011contour} and
MCG \cite{arbelaez2014multiscale}, but also with some RGB-D based methods, 
\eg gPb+NG \cite{gupta2013perceptual}, SE+NG+ \cite{gupta2014learning},
and COB \cite{maninis2016convolutional}.
The precision-recall curves of boundary and region measures are displayed 
in \figref{fig:img_seg_nyud}.
The numeric comparison is summarized in \tabref{tab:img_seg_nyud}.
Our RCF with VGG16 achieves higher F-measure score than COB on the region 
measure, while performs slightly worse than original COB on the boundary measure.
With ResNet50 as the backbone net, RCF achieves similar performance with
COB on the boundary measure but 1.6\% higher on the region measure.
Moreover, both COB and RCF outperform traditional methods by a large margin,
which demonstrates the importance of accurate edges in the classic image segmentation.

\section{Conclusion}
In this paper, we introduce richer convolutional features (RCF), 
a novel CNN architecture which makes good usage of feature hierarchies 
in CNNs, for edge detection.
RCF encapsulates both semantic and fine detail features by leveraging 
all convolutional features.
RCF is both accurate and efficient, making it promising to be applied 
in other vision tasks. 
We also achieve competitive results when applying RCF edges 
for classical image segmentation.
RCF architecture can be seen as a development direction of fully convolutional
networks, like FCN \cite{long2015fully} and HED \cite{xie2017holistically}.
It would be interesting to explore the effectiveness of our network architecture 
in other hot topics \cite{zhao2018hi,liu2017rsrn,zeng2017people,xu2018surface}.
Source code is available at \url{https://mmcheng.net/rcfedge/}.

\newcommand{\gPbUCM}{gPb-UCM \cite{arbelaez2011contour}}
\newcommand{\MCG}{MCG \cite{arbelaez2014multiscale}}
\newcommand{\gPbNG}{gPb+NG \cite{gupta2013perceptual}}
\newcommand{\SENG}{SE+NG+ \cite{gupta2014learning}}
\newcommand{\COB}{COB \cite{maninis2016convolutional}}
\newcommand{\RCFResNet}{RCF-ResNet50}
\begin{figure}[!ht]
    \centering
    \begin{tabular}{c||c|c|c|c} \hline
    	\multirow{2}*{Methods} & \multicolumn{2}{c|}{Boundaries ($F_b$)} 
        	& \multicolumn{2}{c}{Regions ($F_{op}$)} \\ \cline{2-5}
        	& ODS & OIS & ODS & OIS \\ \hline
        \gPbUCM    & 0.631 & 0.661 & 0.242 & 0.283 \\
        \MCG       & 0.651 & 0.681 & 0.264 & 0.300 \\
        \gPbNG     & 0.687 & 0.716 & 0.286 & 0.324 \\
        \SENG      & 0.706 & 0.734 & 0.319 & 0.359 \\
        \COB       & \textbf{0.783} & \textbf{0.804} & 0.353 & 0.396 \\ \hline
        RCF        & 0.775 & 0.798 & 0.364 & \textbf{0.409} \\ 
        \RCFResNet & 0.782 & 0.803 & \textbf{0.369} & 0.406 \\ \hline
    \end{tabular}
    \caption{Evaluation results of boundaries ($F_b$ \cite{martin2004learning})
    and regions ($F_{op}$ \cite{pont2016supervised}) on the 
    NYUD test set \cite{silberman2012indoor}.}
    \label{tab:img_seg_nyud}
\end{figure}

\section*{Acknowledgments}
This research was supported by NSFC (NO. 61620106008, 61572264),
the national youth talent support program,
Tianjin Natural Science Foundation for Distinguished Young Scholars (NO. 17JCJQJC43700),
Huawei Innovation Research Program.

\bibliographystyle{IEEEtran}
\bibliography{edge}

\end{document}